\documentclass[letterpaper, 10 pt, conference]{ieeeconf}  
\usepackage{amsmath}
\usepackage{algorithm}
\usepackage{algorithmic}
\usepackage{mathrsfs}
\usepackage{amssymb}
\usepackage{graphicx}
\usepackage[bookmarks=false]{hyperref}

\usepackage{pifont}

\IEEEoverridecommandlockouts   

\title{\LARGE \bf
Mechatronic Design of a Dribbling System for RoboCup Small Size Robot
}

\author{Zheyuan Huang$^{1}$, Yunkai Wang$^{1}$, Lingyun Chen$^{1}$, Jiacheng Li$^{1}$, Zexi Chen$^{1}$ and Rong Xiong$^{1}$
\thanks{$^{1}$Zhejiang University, Zheda Road No.38, Hangzhou, Zhejiang Province, P.R.China
{\tt\small rxiong@iipc.zju.edu.cn}}%
}

\begin{document}

\maketitle
\thispagestyle{empty}
\pagestyle{empty}

\begin{abstract}

RoboCup SSL is an excellent platform for researching artificial intelligence and robotics. The dribbling system is an essential issue, which is the main part for completing advanced soccer skills such as trapping and dribbling. In this paper, we designed a new dribbling system for SSL robots, including mechatronics design and control algorithms. For the mechatronics design, we analysed and exposed the 3-touch-point model with the simulation in ADAMS. In the motor controller algorithm, we use reinforcement learning to control the torque output. Finally we verified the results on the robot.

\end{abstract}

\section{INTRODUCTION}

The Small Size League(SSL)\cite{robocup}, as one of the most famous and grand robotic competitions in the RoboCup\cite{ssl}, has a history of more than 20 years. As an important part of the soccer robot, the dribbling system can achieve advanced actions such as trapping and dribbling, which is crucial in football strategies. The dribbling system has therefore been innovated and improved by various teams. The ``ZJUNlict''\cite{zju_champion_2013,zju_champion_2014} is a SSL team with fifteen years of history from Zhejiang University and we got the champion of Division A in RoboCup 2018\cite{zju_2018}. The crucial part of our improvements is the new dribbling system with mechatronic design and control algorithms. In the motor controller algorithm, we used reinforcement learning to control the torque output of the motor in the current complex environment and verify the results on the robot.

\section{RELATED WORK}

In wheeled soccer robot competitions of RoboCup, research on improving the ability to dribbling has been ongoing.\par
Chikoshi et al. \cite{rw-1} got the forward and inverse kinematics equation between the ball-motion and two active wheels, and used experiments to evaluate this method. Hoogendijk et al. \cite{rw-2} measured frequency response functions of the prototype mechanism, and designed and tested a PI feedback controller. Li et al. \cite{rw-3} used nonlinear model predictive control(NMPC) method to fulfill the dribbling control task. They let the robot dribble the ball and move along the eight-shaped reference path, then evaluated the difference between the real and ideal position of the ball. Yoshimoto et al. \cite{rw-4} designed an auto centering roller with screw thread. It can increase the amount of force which it can apply to the ball, and push the ball to the middle of it as well.

In our opinion, the dribbling system should have at least two capabilities, cushioning capacity and holding ability. Since the ability of holding ball is a system consisting of the three parts of the dribbler, the ball and the carpet, which has high dynamics, more research is carried out on the static situation of the robot. There are few studies on the ability of dribbling in the case of high-speed movement and the ability of cushioning.

\section{Dribbler Design}
\begin{figure}[h!t]
    \centering
    \begin{minipage}{.23\textwidth}
        \centering
        \includegraphics[height=1.2in]{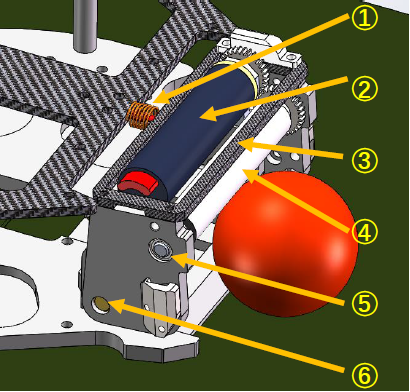}
        \caption{Typical Dribbler. \ding{172}.Unidirectional Damper \ding{173}.Dribble Motor \ding{174}.Connect Shelf \ding{175}.Dribbling-bar \ding{176}.Side Plate \ding{177}.Rotational Joint}
        \label{dribbler}
    \end{minipage}%
    \begin{minipage}{.23\textwidth}
        \centering
        \includegraphics[height=1.2in]{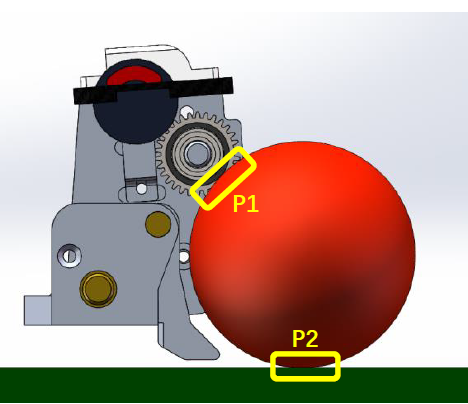}
        \caption{2-touch-point Model}
        \label{2-touch-point}
    \end{minipage}
\end{figure}

The SSL robots do not really have foot like human beings. Instead, they have dribblers. A dribbler is a device that can help dribble and catch the ball and typically it can be devided into 2 main parts --- the mechanical part and the motor control part. 
As shown in Fig. \ref{dribbler}, the mechanical part of a typical dribbler has the following features. A shelf connects 2 side plates and the dribbling motor is fixed on one side plate. Between the 2 side plates is a cylindrical dribbling-bar driven by the dribbling motor. The whole device has only one degree of freedom of rotation and the joints are fixed on the robot flame. Usually there is a unidirectional spring-damping system locates between the shelf and the robot frame to help improve the stability of dribbling as well as absorbing the energy when catching the ball. The dribbling-bar driven by the dribbling motor provides torque to make the ball spin backward when the contact between the ball and dribbling-bar exits so that the ball can be ‘locked’ by this device in ideal conditions. And the carpet provides supporting force and frictional force and therefore there are 2 touch points on the ball and in this paper we called it a 2-touch-point model (Fig. \ref{2-touch-point}).

For the motor control part, most teams try to keep the dribbling-bar at a constant rotational speed when dribbling the ball and therefore it is actually an open loop control mode for dribbling and the dribbling performance is mainly determined by the spring-damping system of the dribbling device. Unfortunately, this 2-touch-point dribbler with unidirectional spring-damping system and passive control mode does not provide ideal dribbling performances. It is quite easy for the ball to bounce back and forth when launching the dribbling motor. The device might also not absorb enough kinetic energy of the moving ball when catching it so it will bounce back and there occurs a catching failure. Actually it is also hard to greatly improve its performance by simply changing the material of dribbling-bar, adjusting the damping and stiffness of the spring-damping system or adjusting the rotational speed of motor. This structure has natural defects with passive control mode.

\begin{figure}[h!t]
    \centering
    \begin{minipage}{.23\textwidth}
      \centering
      \includegraphics[height=1.2in]{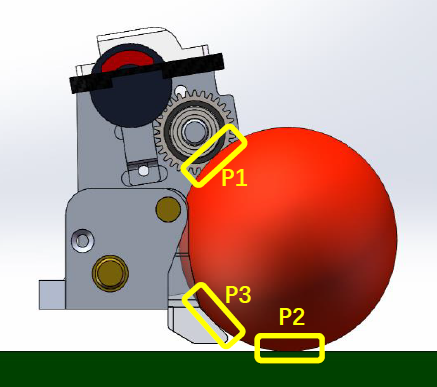}
      \caption{3-touch-point Model}
      \label{3-touch-point}
    \end{minipage}%
    \begin{minipage}{.24\textwidth}
      \centering
      \includegraphics[height=1.2in]{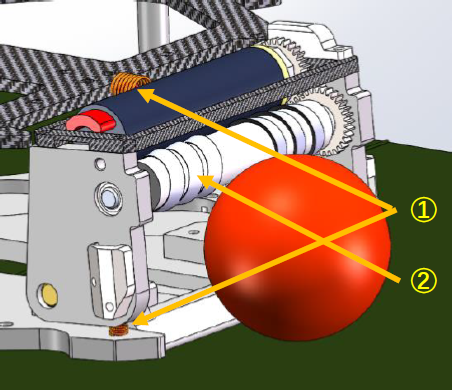}
      \caption{New Damper \ding{172}.Bidirectional Damper \ding{173}.Screw Dribbling Bar}
      \label{new-damper}
    \end{minipage}
\end{figure}

In order to overcome the problems above, we devoted ourselves on the dribbler. Firstly, we adjust the geometry parameters of the device so that the ball can touch the chip shovel in steady state, which means both carpet and chip shovel can provide supporting force and frictional force to the ball so we called it a 3-touch-point model (Fig. \ref{3-touch-point}). Hopefully this design will limit the bouncing space and it will be much easier for the coming ball to enter a steady state. In addition, we found that there will be a hard contact between the side plates and the baseplate when the dribbler hits the baseplate. So besides the foam between the shelf and the robot frame, we stick 1.5 mm thick tape between the side plates and baseplate so there will be a soft contact when the dribbler hits the baseplate. Actually this design makes up a bidirectional spring-damping system (Fig. \ref{new-damper}) and improves the dynamic behavior of the dribbler. Hopefully it can reduce the bouncing amplitude of the ball when dribbling as well as absorbing more kinetic energy when catching the ball. To improve the dribbling performance when the robot rotates or moves laterally, we also made a dribbling-bar with screw using 3D printing rubber so that it can provide lateral force to the ball when dribbling as shown in Fig.\ref{new-damper}.  

\begin{figure}[h!t]
    \centering
    \includegraphics[width=3.0in]{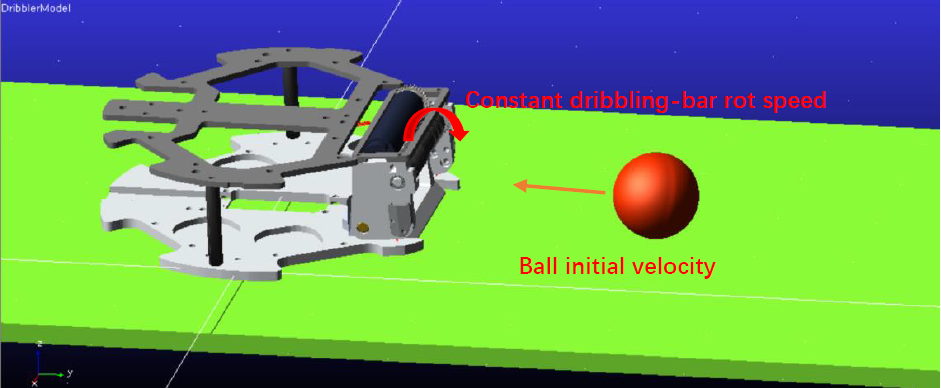}
    \caption{Simulation Environment in ADAMS}
    \label{ADAMS}
\end{figure}

\begin{figure}[h!t]
    \centering
    \includegraphics[width=3.2in]{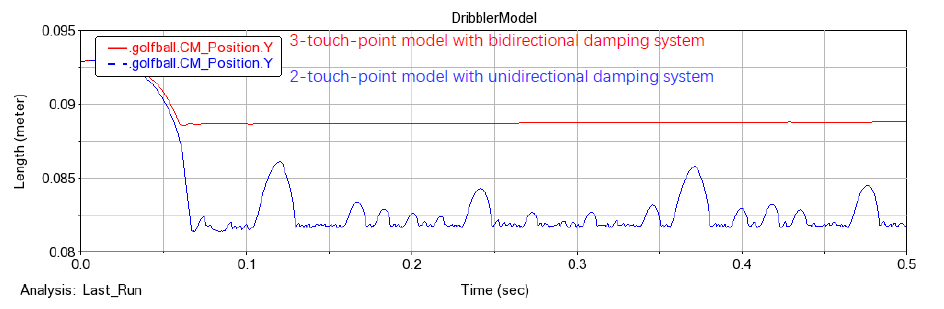}
    \caption{Results of Ball Position Comparison}
    \label{ball-position}
\end{figure}

\begin{figure}[h!t]
    \centering
    \includegraphics[width=3in]{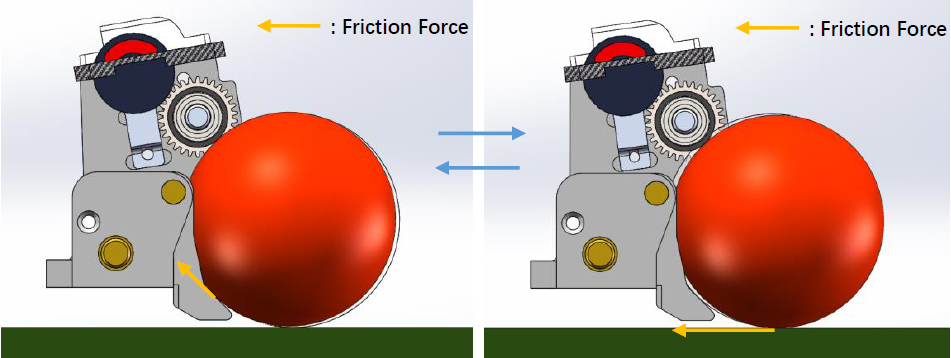}
    \caption{Dynamic Steady State of 3-touch-point Model}
    \label{D-3-touch-point}
\end{figure}

\begin{figure}[h!t]
    \centering
    \includegraphics[width=3in]{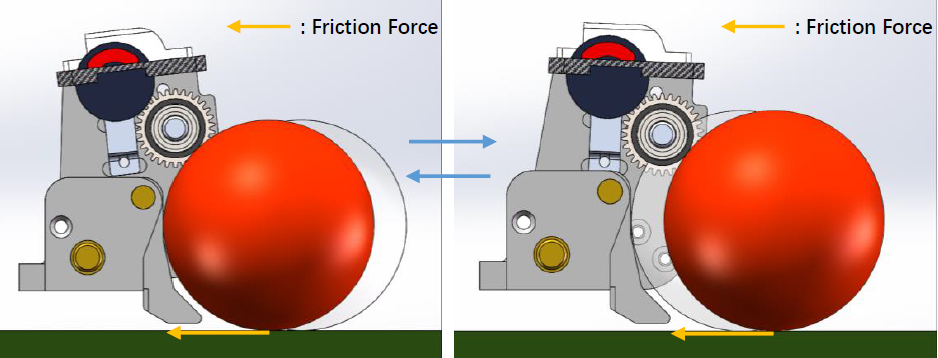}
    \caption{Dynamic Steady State of 2-touch-point Model}
    \label{D-2-touch-point}
\end{figure}

In order to explore the mechanism of the improvements, simulation models of 3-touch-point model with bidirectional spring-damping system compared with the 2-touch-point model with unidirectional spring-damping system were built in ADAMS.
A constant rotation speed of dribbling-bar with $3300r/min$ was given and the ball was released with initial speed of $0.1m/s$ to hit the dribbler (Fig. \ref{ADAMS}). From the simulation results of ball positions (Fig. \ref{ball-position}), the dribbling of the 3-touch-point model was significantly more stable than that of the 2-touch-point-model. It could also be seen that there is no strict steady state, the ball will keep bouncing back and forth and we judge the steady state by the bouncing amplitude which means, if the bouncing amplitude is small enough that the ball never bounces off the dribbler, we can judge it as a steady state. The result also explained why a 3-touch-point model is better than a 2-touch-point model. Normally the dynamic friction coefficient between the ball-carpet surface is greater than that of the ball-chip shove surface. Therefore, when the ball driven by the dribbling-bar moves from the carpet on to the chip shove surface, there will be a sudden drop of frictional force, and the ball will be pushed back on the carpet. And once the ball touches the carpet, there will be a sudden increase of frictional force, the ball will be driven onto the chip shove again. In this kind of state, the amount of spring compression will not change much so that the dribbling system will enter a periodical dynamic steady state (Fig. \ref{D-3-touch-point}). In contrast, with a 2-touch-point system, the friction force will not change much so the ball will enter much more into the dribbler and there will be a bigger compression of the spring-damping system. Therefore the ball will also be pushed back more and totally the bouncing amplitude will be much greater, or even the ball will bounce off the dribbler(Fig. \ref{D-2-touch-point}).

  
With the structural innovation above, we create a quite good passive control dribbler. But considering the real competition environment, the condition will not be that idealistic and more complex movements are needed, indeed. For example, when two robots scramble for a ball, we want our robot able to turn around while dribbling so that it can make space for passing. Also when all shot space is blocked by defenders we want our robot able to do some actions like moving laterally while dribbling to create space to score. In a word, a stronger dribbler is urgently in need. Active controller design based on deep reinforcement learning(DRL) is developed.

\section{Active Controller Based on DRL}

\begin{figure}
    \centering
    \includegraphics[height=1.28in]{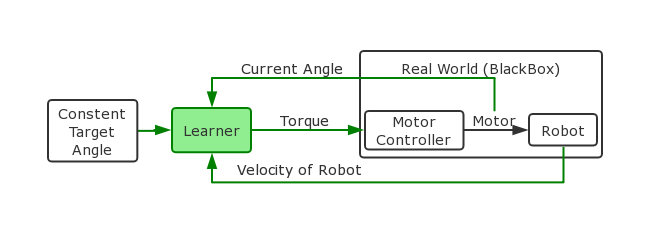}
    \caption{Basic schema of the Active Controller}
    \label{controller_structure}
\end{figure}

The schema of the active controller for the dribbler is shown in Fig. \ref{controller_structure}. From the simulation results of the mechanical system, we found that the control effect is positively correlated with the torque under the three-point ball control model regardless of whether it is in a stationary state or with a ball moving state. However, considering the heat generation problem of the motor, On the one hand, the motor generates heat during operation, and on the other hand, the air dissipates heat. Therefore, in the case of continuous high torque operation, the motor will heat up or even burn out. In response to such a complex situation, we abandoned the traditional control algorithm and used the DRL algorithm to build the controller.\par
The input to the controller is a dribbler optimal angle calculated in the mechanical simulation. For this DRL controller, the torque closed-loop control of the motor and the interaction of the robot with the real world are unknown. Since it is necessary to dribble the ball in different directions, in addition to inputting the current angle into the DRL Controller, it is also necessary to input the current robot speed at the same time.\par
The control method consists of two parts: a traditional control algorithm for torque and a DRL control for angles, which are both explained next.

\subsection{Torque Control for Dribbler Motor}

The torque control model for dribbler motor is based on team Skuba's design \cite{skuba_torque_control,skuba_2011,skuba_hardware}. The motor's dynamic equation is derived from Maxon motor data \cite{maxon_motor_data}.

\begin{equation}
\tau_m=(\displaystyle\frac{k_m}{R})\cdot u - (\displaystyle\frac{k_m}{R\cdot k_n})\cdot \omega
\label{Elec_E1}
\end{equation}
where, $\tau_m$ is the output torque of the motor, $k_m$ is the torque constant of the motor, $R$ is the resistance of the motor, $u$ is the input voltage, $k_n$ is the motor speed constant, $\omega$ is the angular velocity of the motor. Equation \ref{Elec_E1} shows the relationship between output torque $\tau_m$, input voltage $u$ and angular velocity $\omega$. The diagram for the torque controller is shown in the Fig. \ref{Elec_P3} below. The angular velocity $\omega_{raw}$ is acquired through the hall sensor. A low pass filter is used to reduce the noise in the velocity measurement. The desired torque $\tau_d$ is set through the simulation result of the ADAMS model.

\begin{figure}
    \centering
    \includegraphics[height=0.8in]{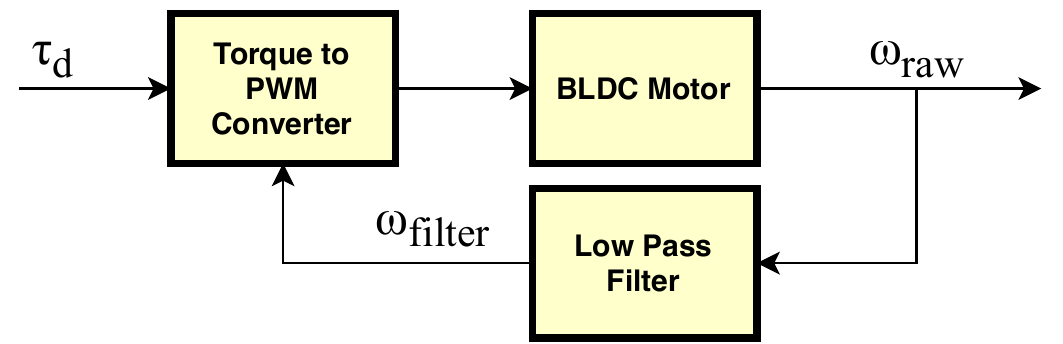}
    \caption{Torque Controller}
    \label{Elec_P3}
\end{figure}

\subsection{Angel Control Using DRL}
In order to allow our robot to interact with the environment to learn the best strategy to control the motor and gradually improve our strategy through trial and error, we used DRL algorithm to achieve better dribbling performance in the simulation system.
\subsubsection{Simulation System}
\begin{figure}[h!t]
  \centering
  \includegraphics[width=2.0in]{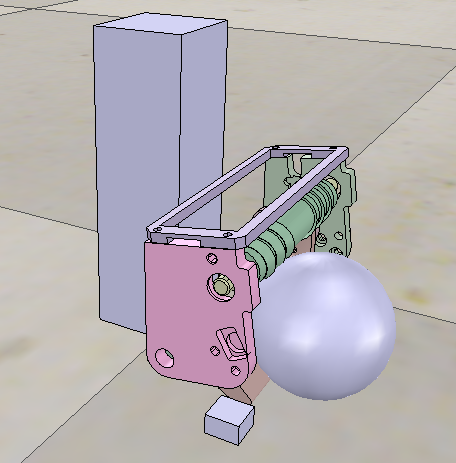}
  \caption{Dribbling System Simulation in V-REP}
  \label{sim}
\end{figure}

We established our dribbling system simulation (\href{https://github.com/ZJUSSL/DribbleSim}{here} to download) in V-REP\cite{V-REP} with bullet\cite{bullet} physical simulation engine according to the actual robot measured parameters, as shown in Fig. \ref{sim}. At the beginning, the ball has a initial speed towards the dribbler. We can get the angle of the dribbling system offset(in the real word we can just use a  proper sensor to get it) as feedback and the dribbler outputs a certain torque to dribble the ball.
\subsubsection{DRL Algorithm}
Twin Delayed DDPG (TD3)\cite{TD3} is one of the state-of-the-art algorithms in DRL, and suitable for our experiment. So in our simulation experiment, the output of the DRL agent is the torque, and the input of the agent is the offset angle of the dribbling system. The pseudo code of TD3 algorithm shown in Algorithm \ref{TD3-alg} and hyperparameters shown in Table \ref{hyperparameters}.
\begin{algorithm}
\caption{TD3\cite{TD3}}
\label{TD3-alg}
\begin{algorithmic}
\STATE Initialize critic networks $Q_{\theta_1}$,$Q_{\theta_2}$, and actor network$\pi_{\phi}$ with random parameters $\theta_1$, $\theta_2$, $\phi$
\STATE Initialize target networks $\theta_{1}^{\prime} \leftarrow \theta_{1}$, $\theta_{2}^{\prime} \leftarrow \theta_{2}$, $\phi^{\prime} \leftarrow \phi$
\STATE Initialize replay buffer $\mathcal{B}$
\FOR{$t=1$ to $T$}
\STATE Select action with exploration noise $a \sim \pi_{\phi}(s) + \epsilon$
\STATE $\epsilon \sim \mathcal{N}(0, \sigma)$  and observe reward $r$ and new state $s^{\prime}$
\STATE Store transition tuple $(s, a, r, s^{\prime})$ in $\mathcal{B}$
\STATE Sample mini-batch of N transitions $(s, a, r, s^{\prime})$ from $\mathcal{B}$
\STATE $\tilde{a} \leftarrow \pi_{\phi^{\prime}}(s^{\prime}) + \epsilon$, $\epsilon \sim clip(\mathcal{N}(0, \tilde{\sigma}), -c, c)$
\STATE $y \leftarrow r + \gamma min_{i=1,2} Q_{\theta_i^{\prime}}(s^{\prime}, \tilde{a})$
\STATE Update critics $\theta_i \leftarrow argmin_{\theta_i}N^{-1} \Sigma{(y-Q_{\theta_i}(s,a))^2}$
\IF{$t$ mod d}
\STATE Update $\phi$ by the deterministic policy gradient:
\STATE $\nabla_{\phi}J(\phi) = N^{-1} \Sigma \nabla_a Q_{\theta_1}(s,a) \arrowvert_{a=\pi_{\phi}(s)}\nabla_{\phi}\pi_{\phi}(s)$
\STATE Update target networks:
\STATE $\theta_{i}^{\prime} \leftarrow \tau \theta_{i} + (1 - \tau)\theta_{i}^{\prime}$
\STATE $\phi^{\prime} \leftarrow \tau \phi + (1 - \tau)\phi^{\prime}$
\ENDIF 
\ENDFOR
\end{algorithmic}
\end{algorithm}

\begin{table}
\centering
\caption{Hyperparameters of TD3}
\label{hyperparameters}
{
\begin{tabular}{ c | c }
\hline
Hidden layer nodes of Actor& (400,300) \\
\hline
Hidden layer nodes of Critic & (400,300) \\
\hline
Activation function & ReLU \\
\hline
Optimizer & Adam \\
\hline
Learning rate & 0.001 \\
\hline
Batch size & 128 \\
\hline
Discount & 0.99 \\
\hline
Soft update rate & 0.005 \\
\hline
Plicy noise & 0.2 \\
\hline
Noise clip & 0.5 \\
\hline
Policy update frequency & 2\\
\hline
\end{tabular}
}
\end{table}

\subsubsection{Reward Setting}
In our simulation system, if the output torque is too small, the dribbler will lose the ball, and if the output torque is too big, the ball will vibrate between the dribbler and the ground and cause the motor overheating. Combine the above points, we designed a reward function in Eqution \ref{r_torque}-\ref{reward-function}
\begin{equation} \label{r_torque}
r_{torque}=\left\{
\begin{aligned}
& 0, & torque \leq 0.02 (mN\cdot m) \\
& -20(torque-0.02), & torque > 0.02 (mN\cdot m)
\end{aligned}
\right.
\end{equation}

\begin{equation}\label{r_offset}
r_{offset} = 2e^{-offset^2/0.004}
\end{equation}

\begin{equation} \label{r_ball}
r_{ball}=\left\{
\begin{aligned}
& 0, & if\ ball\ in\ dribbler \\
& -100, & if\ lose\ ball
\end{aligned}
\right.
\end{equation}

\begin{equation}\label{reward-function}
r = r_{torque} + r_{offset} + r_{ball}
\end{equation}
In Eqution \ref{r_torque}, $0.02(mN\cdot m)$ is the torque threshold that dribbler can hold the ball, $-20$ is the gain of this part. If the torque is too big, this reward will be negative value. In Euqtion \ref{r_offset}, $offset$(rad) is the offset of the dribbling system, $2$ is a gain and $0.004$ controls the smoothness of this part of the reward function. In Eqution \ref{r_ball}, $-100$ is a large penalty value.

\subsubsection{Simulation Result}
\begin{figure}[h!t]
\centering
\begin{minipage}{.25\textwidth}
  \centering
  \includegraphics[height=1.3in]{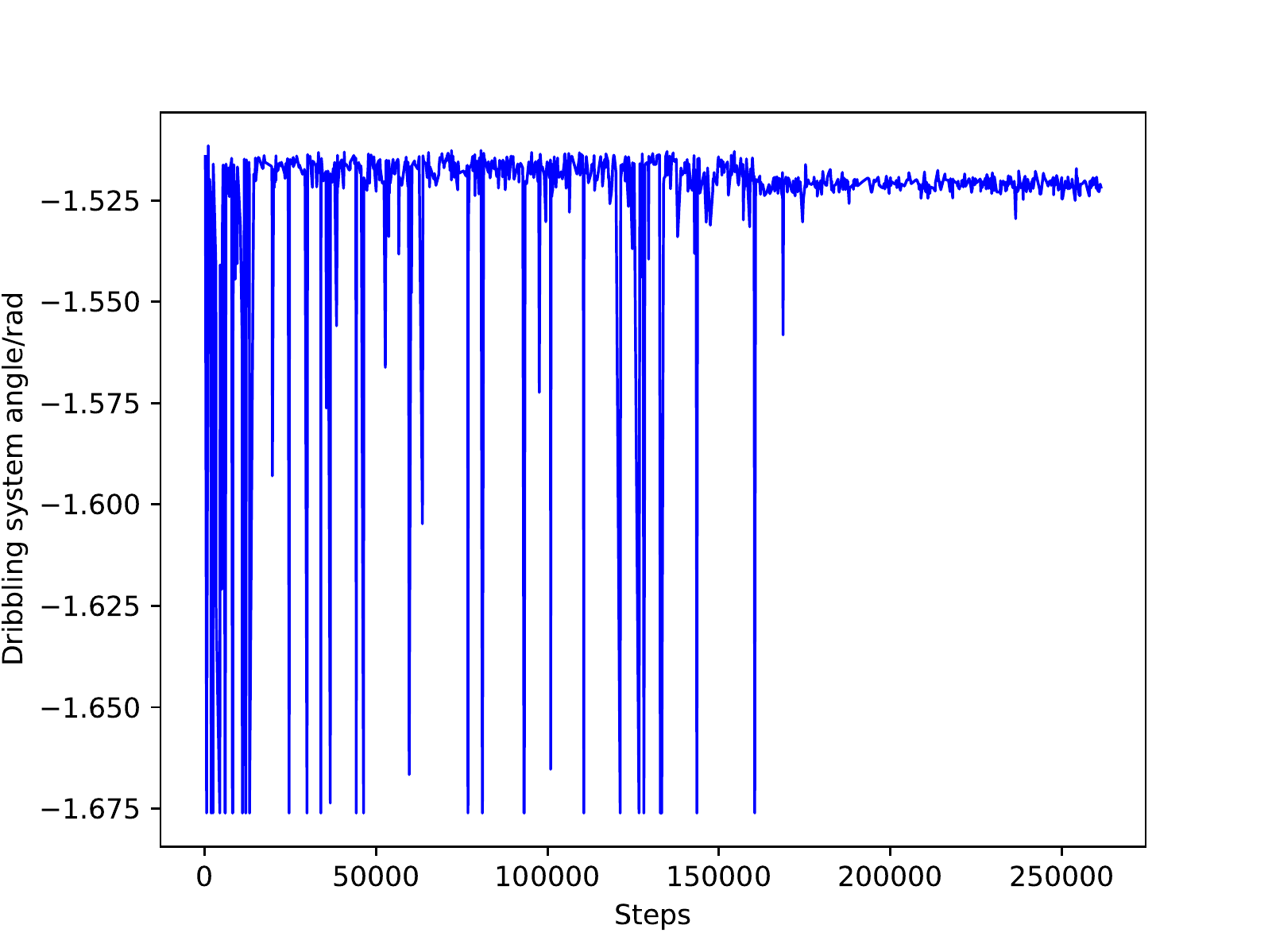}
  \caption{Dribbling System Angle Result}
  \label{angle}
\end{minipage}%
\begin{minipage}{.25\textwidth}
  \centering
  \includegraphics[height=1.3in]{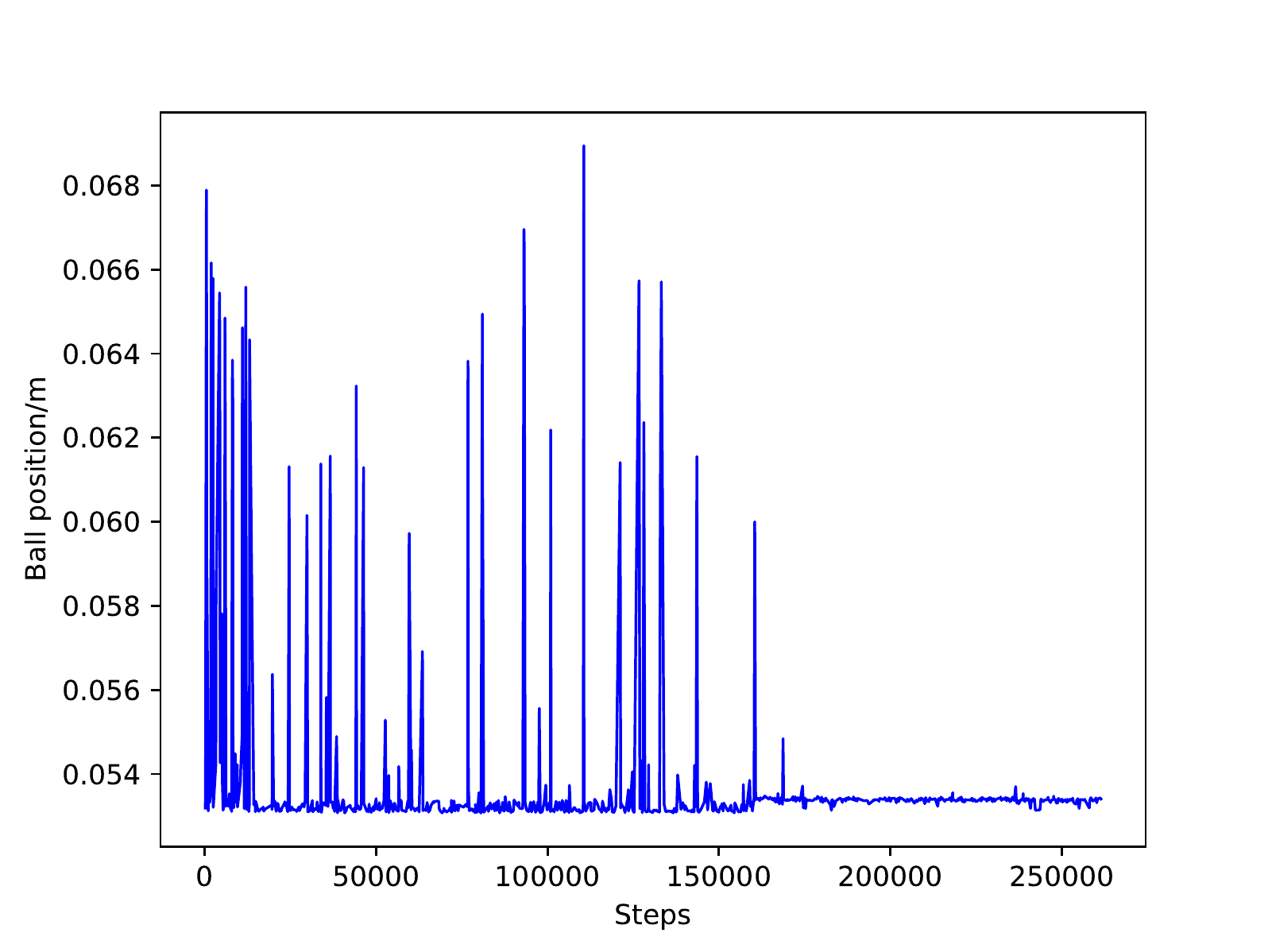}
  \caption{Ball Position Result}
  \label{dist}
\end{minipage}
\end{figure}

\begin{figure}[h!t]
\centering
\begin{minipage}{.25\textwidth}
  \centering
  \includegraphics[height=1.3in]{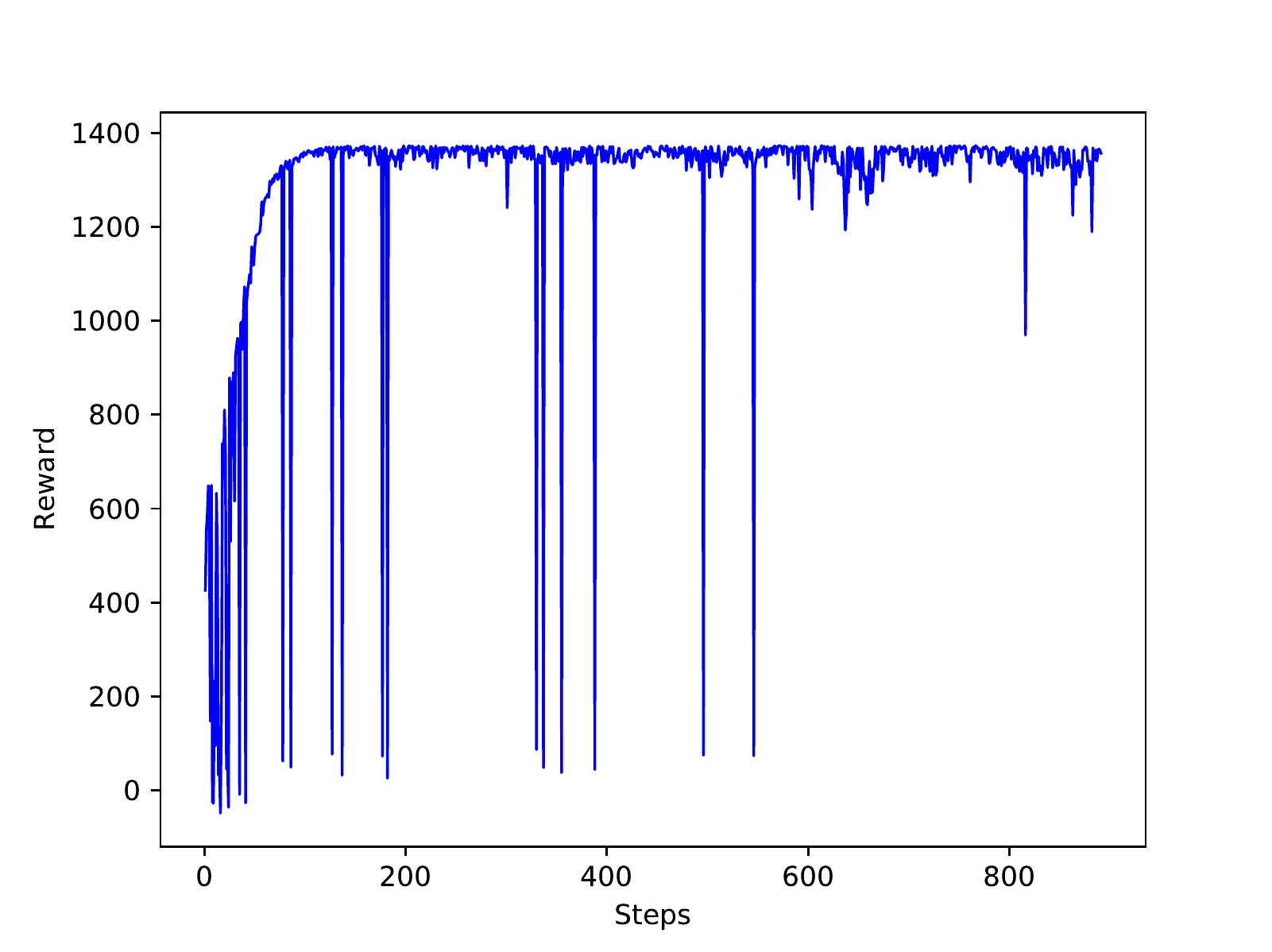}
  \caption{Reward Result}
  \label{reward}
\end{minipage}%
\begin{minipage}{.25\textwidth}
  \centering
  \includegraphics[height=1.3in]{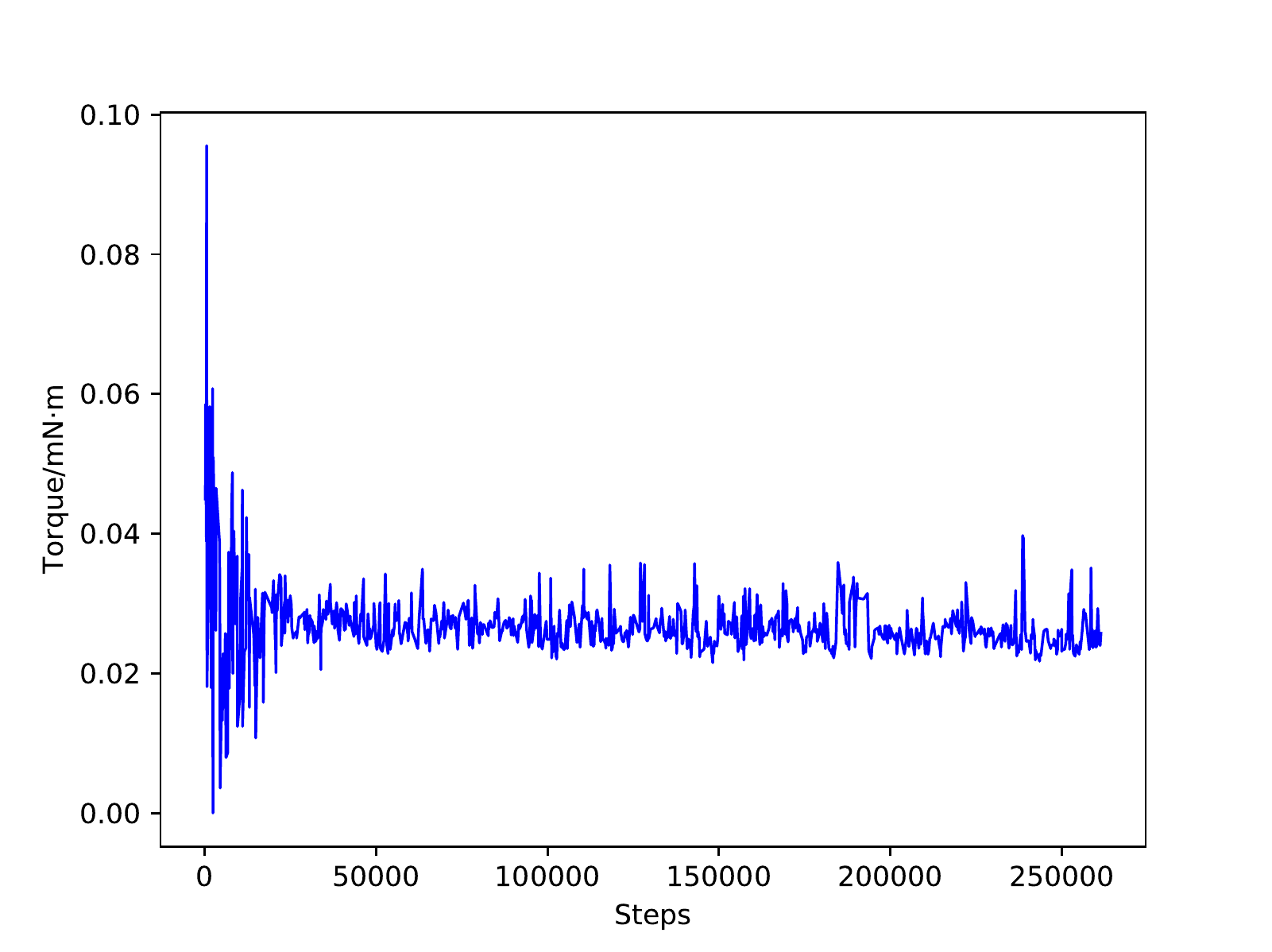}

  \caption{Torque Result}
  \label{torque}
\end{minipage}
\end{figure}

We have trained our TD3 algorithm on Intel i7-7700 CPU and NVIDIA GTX1060 GPU for about 9 hours, the resaults shown in Fig. \ref{angle}-\ref{torque}. In Fig. \ref{angle}, the value of dribbling system angle finally converges to $-1.52 rad$, that is, the dribbling system leans back slightly. In Fig. \ref{dist} shows the position of the ball with respect to the absolute coordinate system. This value measures the severity of the ball's vibration in the dribbler. It finally converges to almost zero, that means our trained DRL agent is able to control the ball very well. Fig \ref{reward} and Fig \ref{torque} shows the training episode reward and the output torque, they all have some noise because of randomness of the training process and artificially increased noise for improving the ability to explore. The output torque fluctuates around $0.027 (mN\cdot m)$, a little bigger than the torque threshold $0.02(mN\cdot m)$, that means $0.027 (mN\cdot m)$ is close to the optimal value of the output torque in this condition. We also tried to give the ball a speed to leave the dribbler in the simulation system, but sadlly the DRL agent couldn't respond in time when the ball got speed, and it just kept increasing the output torque to prevent losing the ball.

\section{Experiments and Results}

According to the catching ability tests, the typical 2-touch-point dribbler with unidirectional spring-damping system could catch a ball with coming speed up to $3 m/s$ and the new 3-touch-point dribbler with bidirectional spring-damping system could catch a ball with coming   speed up to $8.5 m/s$. The results were quite clear that the new dribbler has better dribbling and catching ability.
In addition, we made simple tests to see the effect of screw added on the dribbling bar. The dribbling motor was launched and after the dribbling entering the steady state, we made the robot spin around. The rotational acceleration is $20 deg/s^2$ and the rotation speed was recorded at the time the ball left the dribbler. This simple test was carried out 10 times for both smooth dribbling-bar and screw dribbling-bar, which were made by some same material. As show in Tabel.1 below, the average escape speed of smooth dribbling-bar is $402 deg/s$ and for the screw dribbling-bar is $622 deg/s$.
Finally tests to verify the contribution of active control mode were carried out. this time after the dribbling entering the steady state, we made the robot move back. The backward acceleration is $ 0.2m/s^2$ and the backward speed was recorded at the time the ball left the dribbler. This test was also carried out 10 times for both active control mode and passive control mode. As show in Tabel.2 below, the average escape speed of active control mode is $ 2.5m/s$ and for the passive control mode is $ 1.63m/s$. So it was proved that the active control mode significantly improve the dynamic dribbling performance of dribbler.

\begin{table}
\centering
\caption{Dynamic Dribbling Ability Comparison Between Smooth Dribbling-bar and Screw Dribbling-bar}
\label{tab1}
\resizebox{.46\textwidth}{!}{%
\begin{tabular}{| c | c | c | c | c | c | c | c | c | c | c | c |}
\hline
Dribbling-bar Type & 1 & 2 & 3 & 4 & 5 & 6 & 7 & 8 & 9 & 10 & Average\\
\hline
Smooth Dribbling-bar $(deg/s)$ & 400 & 340 & 380 & 360 & 380 & 420 & 400 & 420 & 400 & 520 & 402\\
\hline
Screw Dribbling-bar $(deg/s)$ & 600 & 580 & 580 & 580 & 620 & 680 & 620 & 680 & 640 & 640 & 622\\
\hline
\end{tabular}
}
\end{table}

\begin{table}
\centering
\caption{Dynamic Dribbling Ability Comparison Between Active Control Mode and Passive Control Mode}
\label{tab2}
\resizebox{.46\textwidth}{!}{%
\begin{tabular}{| c | c | c | c | c | c | c | c | c | c | c | c |}
\hline
Control Mode & 1 & 2 & 3 & 4 & 5 & 6 & 7 & 8 & 9 & 10 & Average\\
\hline
Active $(m/s)$ & 2.40 & 2.50 & 2.40 & 2.60 & 2.50 & 2.50 & 2.50 & 2.80 & 2.30 & 2.50 & 2.50\\
\hline
Passive $(m/s)$ & 1.80 & 1.70 & 1.40 & 1.70 & 1.60 & 1.60 & 1.50 & 1.80 & 1.70 & 1.50 & 1.63\\
\hline
\end{tabular}
}
\end{table}

\section{CONCLUSIONS}

In this paper, the complete dribbling system was demonstrated, including its mechatronic design, simulation analysis and control algorithm implementation. The test results show that the 3-touch-point design can effectively improve the stability of the ball. Although the controller using DRL does not complete the stability control of omnidirectional dribbling with high-speed, it is still a great method for complex nonlinear controlling problems. We will try further on this issue. \href{https://youtu.be/P6ZVcEb5dNs}{Here} you can find our video of test results.

\section*{ACKNOWLEDGMENT}
This work was supported by the National Nature Science Foundation of China under Grant U1609210 and  Science and Technology Project of Zhejiang Province under Grant 2019C01043.

\end{document}